\documentclass[10pt,twocolumn,letterpaper]{article}
\usepackage{color}
\usepackage[table]{xcolor}
\usepackage{cvpr}
\usepackage{times}
\usepackage{epsfig}
\usepackage{graphicx}
\usepackage{dcolumn}
\usepackage{fixltx2e}
\usepackage{amsmath}
\usepackage{amssymb}
\usepackage{xfrac}
\usepackage[font=small]{caption}
\usepackage[font=small]{subfig}
\definecolor{Gray}{gray}{0.9}
\newcolumntype{M}{>{\centering\arraybackslash}m{\dimexpr.19\linewidth-2\tabcolsep}}
\usepackage[pagebackref=true,breaklinks=true,letterpaper=true,colorlinks,bookmarks=false]{hyperref}

\cvprfinalcopy 

\def\cvprPaperID{710} 

\ifcvprfinal\fi
\begin{document}

\title{Reproducible Evaluation of Pan-Tilt-Zoom Tracking}

\author{
    Gengjie Chen\thanks{This work was conducted while Gengjie Chen was doing a MITACS Globalink internship at Polytechnique Montr\'eal}\\
    Dept. Electron. \& Commun. Eng.\\
    Sun Yat-sen University\\
    {\tt\small chgengj@mail2.sysu.edu.cn}
\and
    Pierre-Luc St-Charles, Wassim Bouachir,\\ Thomas Joeisseint, Guillaume-Alexandre Bilodeau\\
    LITIV lab., Dept. Computer \& Software Eng.\\
    Polytechnique Montr\'eal\\
    {\tt\small\{firstname.lastname\}@polymtl.ca}
\and
    Robert Bergevin\\
    LVSN - REPARTI\\
    Universit\'e Laval\\
    {\tt\small robert.bergevin@gel.ulaval.ca}
}

\maketitle

\begin{abstract}
Tracking with a Pan-Tilt-Zoom (PTZ) camera has been a research topic in computer vision for many years. However, it is very difficult to assess the progress that has been made on this topic because there is no standard evaluation methodology. The difficulty in evaluating PTZ tracking algorithms arises from their dynamic nature. In contrast to other forms of tracking, PTZ tracking involves both locating the target in the image and controlling the motors of the camera to aim it so that the target stays in its field of view. This type of tracking can only be performed online. In this paper, we propose a new evaluation framework based on a virtual PTZ camera. With this framework, tracking scenarios do not change for each experiment and we are able to replicate online PTZ camera control and behavior including camera positioning delays, tracker processing delays, and numerical zoom. We tested our evaluation framework with the Camshift tracker to show its viability and to establish baseline results. 
\end{abstract}

\section{Introduction}
\label{intro}

Tracking with a single Pan-Tilt-Zoom (PTZ) camera has been a research topic in computer vision for many years \cite{Roh2007, Babu2007, Bagdanov2006, Darvish2011,Cai2013}. However, it is very difficult to assess the progress that has been made on this topic because there is no standard evaluation methodology. The difficulty in evaluating PTZ tracking arises from its dynamic nature. In contrast to other forms of tracking, PTZ tracking involves both locating the target in the image and controlling the motors of the camera to aim it so that the target stays in its field of view (FOV). This type of tracking can only be performed online. As a result, it is very difficult to compare two algorithms with a real PTZ camera because the same experiment is not repeatable. Even under a strict scenario with actors performing predefined actions, the tracking conditions will never be totally identical. 

Recent datasets like VOT2014\footnote{\url{http://www.votchallenge.net/vot2014/}} only test the quality of the target location in each frame. Although important, it does not account for the online constraint of tracking with a PTZ camera. For example, in an online setting with a PTZ camera, if an algorithm processes a frame in one second, it is essentially blind during this entire time lapse. It means that the target may move over a large distance between two observations. Moreover, centering the camera on its previous location may result in the target leaving the FOV. In general, with PTZ tracking, there is a compromise between two requirements:
\begin{enumerate}
\vspace{-6pt}
\item Designing a fast tracker that can process every frame without dropping any, and that always recenters the camera at the target's previous position (which is a good approximation of its next position since the frame processing rate is high). Such an algorithm is however more likely to localize the target poorly.
\vspace{-6pt}
\item Designing a slower, but more sophisticated tracker that can localize the target accurately. Since being slow also means being blind for long periods of time, in order to improve robustness to fast target motion, another algorithm needs to be designed to control the camera. A typical approach is to determine the target's most probable location in the next frame, and center the FOV on that position.
\end{enumerate}
\vspace{-6pt}

In short, the processing time budget is important in PTZ tracking because of its online nature, and slow processing means missed observations, which might be crucial for accurate results. With this paper, we hope to inspire the development of better PTZ tracking methods by proposing a virtual camera that allows panning, tilting, and zooming inside pre-recorded spherical panoramic videos. Under our new proposed evaluation framework\footnote{\url{http://www.polymtl.ca/litiv/vid/index.php}}, tracking conditions do not change for each experiment. Besides, we replicate online PTZ camera control behavior by considering camera positioning delays, tracker processing delays, and numerical zoom. Note that in this work, we focus only on single object tracking. Our contributions are:
\begin{itemize}
\vspace{-6pt}
\item a publicly available C++ library implementing a virtual PTZ camera that behaves like an actual PTZ camera. It offers basic functionalities (image acquisition, camera movement) as well as online evaluation of tracking performance using four metrics;
\vspace{-6pt}
\item three publicly available spherical panoramic scenarios taken in two real-world environments, featuring a total of $36$ manually annotated tracking sequences for various object types; and
\vspace{-6pt}
\item a set of baseline performance results obtained using the Camshift tracker \cite{bradski1998real} on the abovementioned dataset.
\end{itemize}

\section{Related Work}
To the best of our knowledge, only two works specifically addressed the evaluation of tracking with a PTZ camera \cite{Salvagnini2011,Qureshi2011}. In the work of Qureshi and Terzopoulos \cite{Qureshi2011}, a virtual world is simulated where animated pedestrians can be tracked by various virtual sensors, including virtual PTZ cameras. This approach is very interesting as it allows repeatable evaluation. Its drawback is that it does not reproduce real-world settings such as change in lighting conditions, nor addresses the limits of real camera sensors (resolution, motion blur, etc.) because the scenes are artificial. It was used in the context of sensor networks. Salvagnini et al. \cite{Salvagnini2011} proposed an experimental framework where a real PTZ camera tracks objects moving on a large screen. The goal of this work was the same as ours: it does provide repeatable scenarios for internal use in a given research laboratory, but other research groups cannot repeat the same experiments as they are equipment-specific. Furthermore, the PTZ camera motion is limited to a very small portion of its operating range. 

The evaluation metrics used in previous work on tracking with a PTZ camera are varied but are essentially very similar to those for the evaluation of single object trackers or multiple object trackers. For example, in Cai et al. \cite{Cai2013}, tracking is evaluated with multiple object tracking metrics. There is no specific evaluation of camera control, although for this work the PTZ camera mostly zooms (it does not pan or tilt significantly). In the rest of the literature on this topic however, most authors evaluate camera control to some extent. In both the work of Lee et al. \cite{Lee2012} and Liu et al. \cite{Liu2014} tracking is evaluated by the percentage of frames where the tracked object is in the FOV. Such a metric roughly evaluates both camera control and tracking performance simultaneously. 

Since tracking with a single PTZ camera requires the evaluation of both tracking and camera control performance, Darvish and Bilodeau \cite{Darvish2011} and Salvagnini et al. \cite{Salvagnini2011} proposed metrics for both aspects. Center Location Error (CLE) and overlap ratio \cite{PASCAL2010} were used for tracking accuracy, and the distance between the center of the ground-truth target position and the center of the image was used to evaluate camera control. The assumption for the evaluation of the camera control is that if it is done properly, the target will always be close to the center of the FOV. If not, the probability that the target will leave the FOV after sudden movements or direction changes is high. In addition to those metrics, Darvish and Bilodeau \cite{Darvish2011} and Paillet et al. \cite{Paillet2013} also included a track fragmentation metric, that is, the number of frames for which the target is out of the FOV. 

\section{Evaluation Framework}
Our PTZ evaluation framework is composed of three components: 1) a C++ library that simulates a PTZ camera and includes an evaluator, 2) a collection of spherical panoramic videos for different scenarios, and 3) their corresponding ground-truth annotated sequences. The PTZ simulator grabs panoramic images from a video file, builds the scenario model and provides a typical viewing frustum for the tracker based on camera parameters. The evaluator uses basic ground-truth data and the same camera parameters to generate ground truth bounding boxes for the current FOV, and then compares them with actual tracking results. 

The framework has been designed based on videos captured by a Point Grey Ladybug 3 Spherical camera and OpenGL to project the videos on a sphere. The Ladybug camera gives a near 360$^{\circ}$ spherical view of the scene that can be mapped on such a surface. It is thus possible to design a virtual camera that can observe specific portions of the sphere. Therefore, we obtain a virtual PTZ camera that can be controlled as desired to track objects in pre-recorded videos. For convenience, the center of the spherical model is set at the origin of the world coordinates.

\begin{figure*}
    \centering
    \subfloat[\label{fig:geometry1a}]{\includegraphics[width=0.48\linewidth]{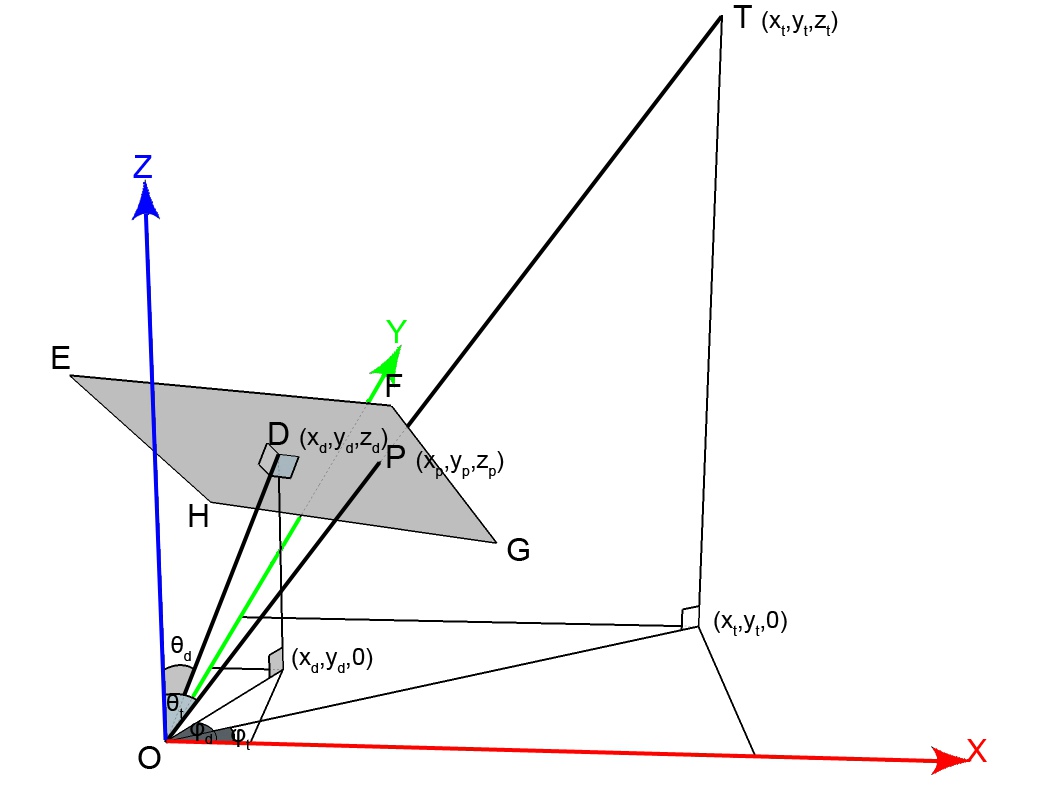}}
    \subfloat[\label{fig:geometry1b}]{\includegraphics[width=0.48\linewidth]{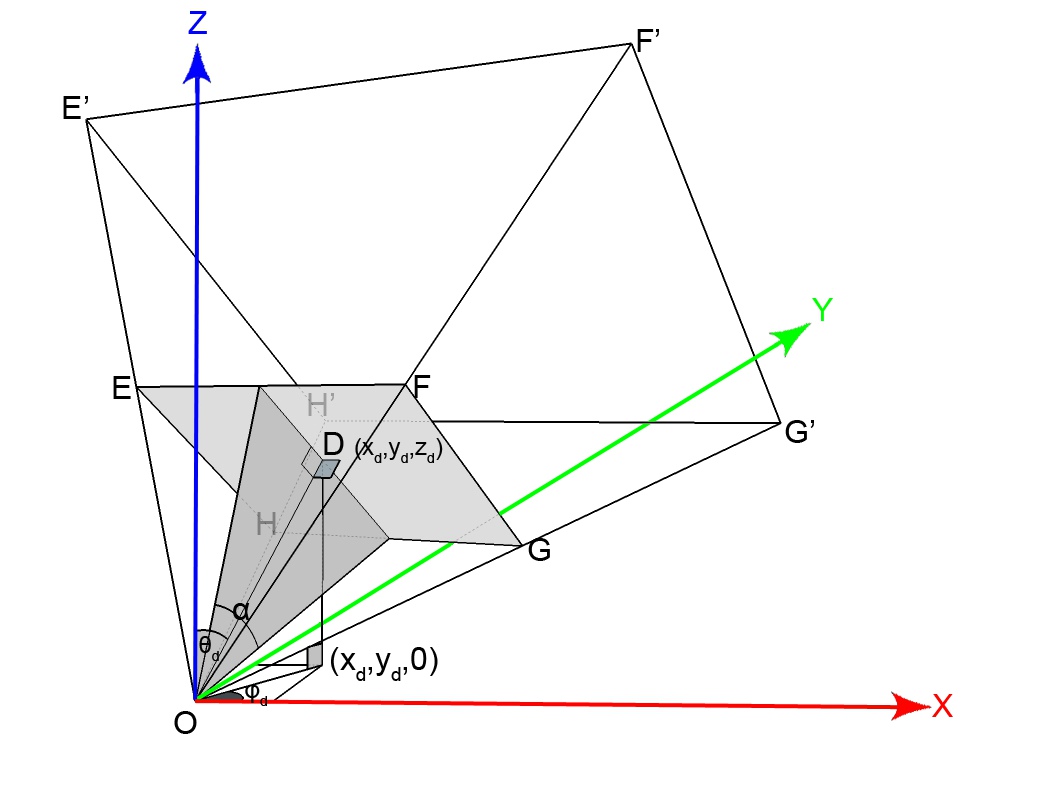}} 
    \caption{Rectangle $EFGH$ is the image plane. Its normal vector $\vec{OD}$ represents the direction vector of the camera. $T$ is the position of the tracked object and its projected point on the image plane is $P$. (a) shows the projection relation in world coordinates, and (b) shows the projection frustum in world coordinates. Frustum $EFGH-E'F'G'H'$ defines the projection volume.}
    \label{fig:geometry1} 
\end{figure*}

\subsection{PTZ Camera Model}
\subsubsection{Camera viewing direction}
After building the model of the scenario, the virtual camera is placed at the origin $O$ as Fig. \ref{fig:geometry1}a shows. Though the position of the camera is constrained, it still has three degrees of freedom. They are: 1) pitching, or forward and backward tilting; 2) yawing, or left and right panning; and 3) rolling, or the rotation on the axis between $O$ and the target point, $T$. Since we are simulating a PTZ camera, rolling is ignored and changing the pitch and yaw angles achieves the functionality of tilting and panning, respectively. Consequently, we use the normal vector $\vec{OD}$ of the image plane to define the direction vector of the camera in world coordinates. This direction is determined by the tilt angle $\theta_d$ and the pan angle $\phi_d$, shown in Fig. \ref{fig:geometry1}a, both of which can be obtained from the position of $D=(x_d, y_d, z_d)$:
\begin{equation}
\theta_d=arccos(\frac{z_d}{\sqrt{x^2_d+y^2_d+z^2_d}})
\end{equation}
\begin{equation}
\phi_d=arctan(\frac{y_d}{x_d})
\end{equation}

Refer to Fig. \ref{fig:geometry1}a for the meaning of the variables.

\subsubsection{Perspective projection}
In the pinhole camera model, the real object is projected onto an image plane through a hole (center of projection). The image plane is located on one side of the pinhole while the object is on the other side. The relation between these two follows the rule of light propagation. It is mathematically equivalent to place the image plane $EFGH$ between the object $T$ and hole $O$, as Fig. \ref{fig:geometry1}a shows. Under this situation, the projected point $P$ is the intersection of image plane $EFGH$ and vector $\vec{OT}$. 

In order to specify the perspective projection, a frustum should be defined as the one shown in Fig. \ref{fig:geometry1}b, noted $EFGH-E'F'G'H'$. All the points inside it will be projected onto the image plane $EFGH$. To determine a frustum, the following parameters are required: 1) the tilt angle $\theta_d$ and pan angle $\phi_d$ of the camera; 2) the vertical FOV $\alpha$; 3) the aspect ratio $r = {\frac{|HG|}{￼|EH|}}$; 4) the distance from the virtual camera to the near clipping plane $n = |OD|$; and 5) the distance from the virtual camera to the far clipping plane.

If we ignore the computer graphics concepts and simply consider the principle of projection, the fourth parameter of the projection model, $n$, can be any positive value that is relatively small, and the fifth parameter (the distance from the viewer to the far clipping plane) is unnecessary. The detailed mathematical pipeline used to obtain the 2D coordinates of a point on the image plane from its 3D coordinates in world space is described next. This ``forward'' pipeline is used to calculate the mapping relation from world coordinates to image coordinates, as opposed to the ``inverse'' pipeline, which calculates the reverse transformation.

\begin{figure*}
    \centering
    \subfloat[\label{fig:geometry2a}]{\includegraphics[width=0.32\linewidth]{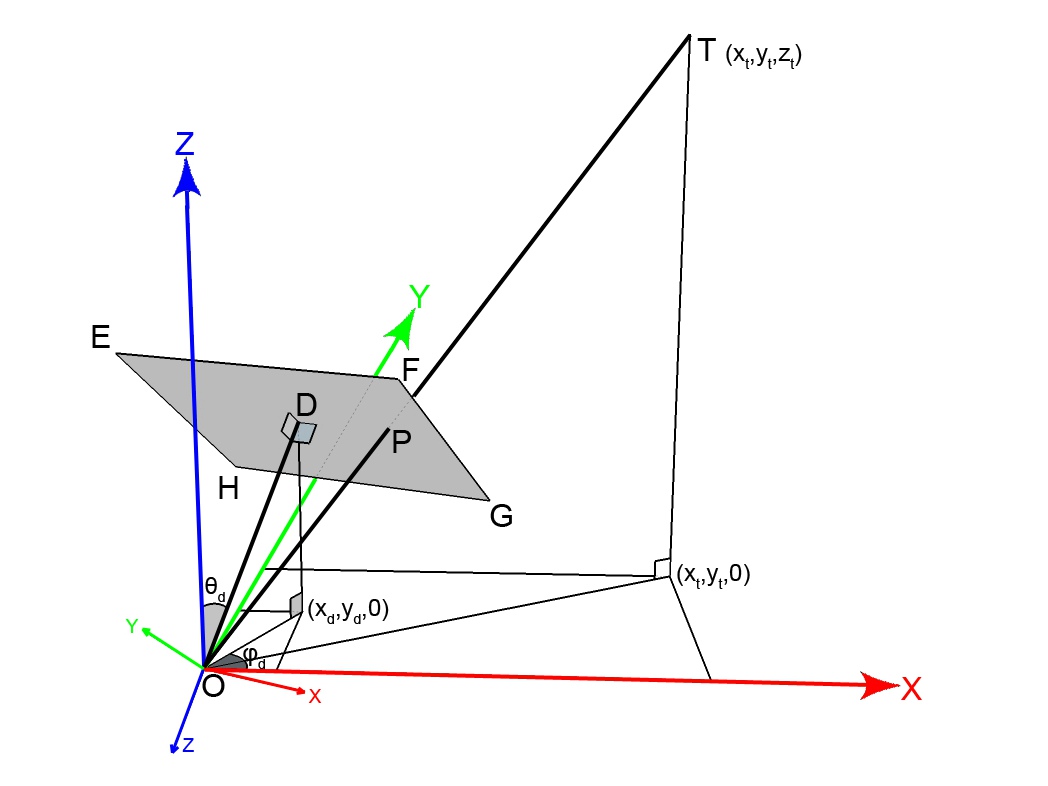}}
    \subfloat[\label{fig:geometry2b}]{\includegraphics[width=0.32\linewidth]{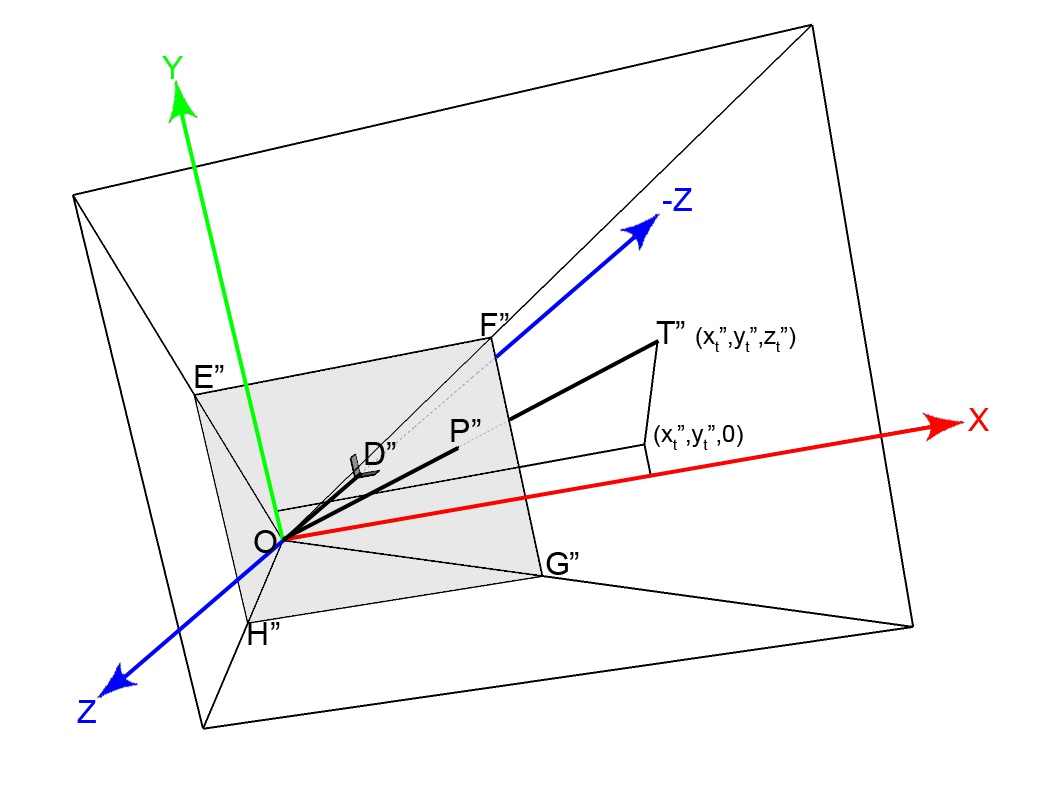}}
    \subfloat[\label{fig:geometry2c}]{\includegraphics[width=0.32\linewidth]{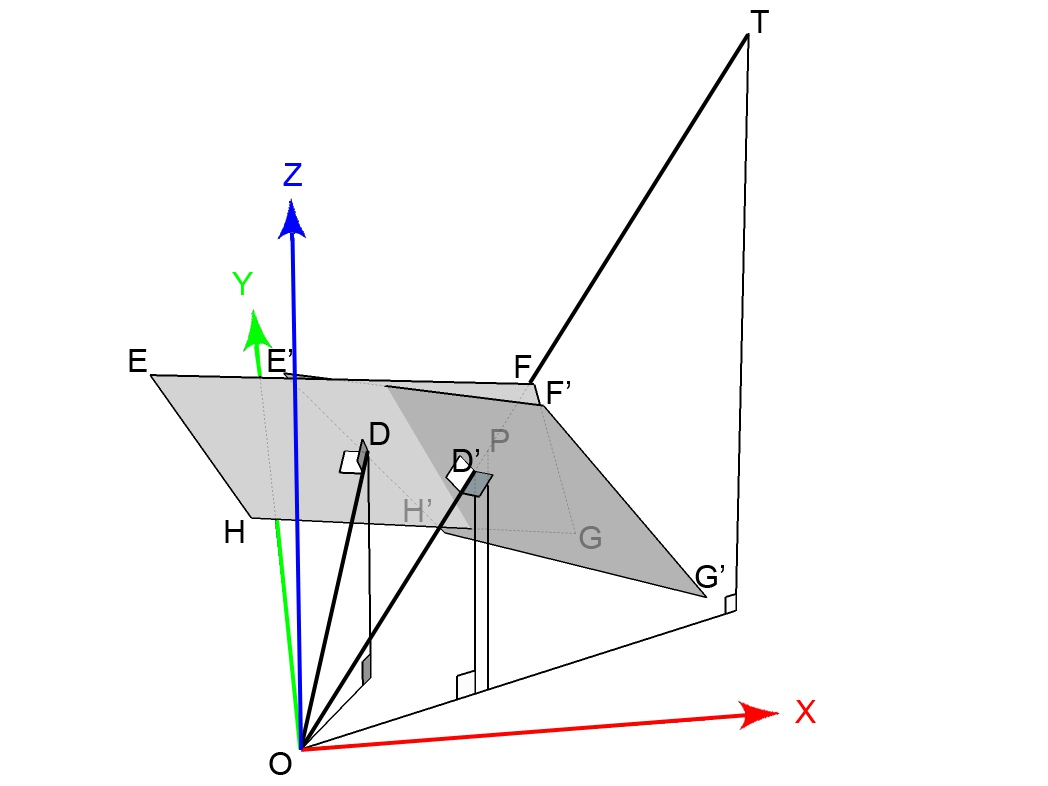}} 
    \caption{Rectangle $EFGH$ is the image plane. Its normal vector $\vec{OD}$ represents the direction vector of the camera. $T$ is the position of the tracked object and its projected point on the image plane is $P$. (a) shows the coordinate system in world space, (b) shows the coordinate system along with the projection volume transformed to camera space (with a similar notation), and (c) shows the camera rotation used to center a point on the image plane. In (c), target $T$ is originally projected to point $P$ on the image plane $EFGH$. Following the rotation, it is projected to the middle $D'$ of image plane $E'F'G'H'$.}
    \label{fig:geometry2} 
\end{figure*}

\subsubsection{From the 3D world to 2D images}
\label{pipeline}
Theoretically, it is possible to compute the transformation that directly projects a 3D point in the model onto the 2D image plane, but the corresponding geometric computations are complex. To simplify the problem, we first transform the coordinate system so the camera points toward the $-Z$ axis and has upward vector $+Y$, as shown in Fig. \ref{fig:geometry2}b. The reason behind using the $-Z$ axis instead of $+Z$ axis is to make the $+X$ axis in right-hand direction and the $+Y$ axis upwards. After this view transformation, the points in 3D space are projected to image coordinates through perspective projection.

\textbf{View transformation} Through view transformation, the coordinates of points in 3D world space are transformed to 3D camera space. In Fig. \ref{fig:geometry2}a, the camera shown in world space has a vertical direction angle $\theta_d$ and a horizontal direction angle $\phi_d$. To make its coordinate system match with the world coordinate system, we use the following transformations: 1) rotate the camera and its target point around the $Z$ axis by an angle of $(\frac{\pi}{2}-\phi_d)$, and 2) rotate the camera and its target point around the $X$ axis by an angle of $(\theta_{d}-\pi)$. Both rotations follow the right-hand rule and the axes are from the world coordinate system.

Mathematically, the rotation which transforms a point $T=(x_t,y_t,z_t)$ in world space to the point $T'' =(x''_t,y''_t,z''_t)$ in camera space is determined by the view transformation matrix $M_{view}$, which is obtained by multiplying the rotation matrices $R_x(\theta_{d}-\pi)$ and $R_z(\frac{\pi}{2}-\phi_d)$. These relations are as follows:

\begin{equation}
\begin{bmatrix}
x''_t\\
y''_t\\
z''_t
\end{bmatrix}=M_{view}{\cdot} \begin{bmatrix}
x_t\\
y_t\\
z_t
\end{bmatrix}=R_x(\theta_{d}{-}\pi){\cdot} R_z(\frac{\pi}{2}-\phi_d){\cdot} \begin{bmatrix}
x_t\\
y_t\\
z_t
\end{bmatrix},
\end{equation}
\noindent where
\begin{equation}
R_x(\theta_{d}-\pi)=\begin{bmatrix}
1 & 0 & 0\\
0 & cos(\theta_{d}-\pi) & -sin(\theta_{d}-\pi)\\
0 & sin(\theta_{d}-\pi) & cos(\theta_{d}-\pi)
\end{bmatrix}
\end{equation}
\noindent and
\begin{equation}
R_z(\frac{\pi}{2}-\phi_d)=\begin{bmatrix}
cos(\frac{\pi}{2}-\phi_d) & -sin(\frac{\pi}{2}-\phi_d) & 0\\
sin(\frac{\pi}{2}-\phi_d)& cos(\frac{\pi}{2}-\phi_d) & 0\\
0 & 0 & 1
\end{bmatrix}
\end{equation}

\textbf{Projection transformation} Through projection transformation, the 3D coordinates of points in the camera space are transformed to the 2D image space. In Fig. \ref{fig:geometry2}b, the target point $T''=(x''_t,y''_t,z''_t)$ is projected to point $P''= (x''_p, y''_p, z''_p)$. According to the relationships between similar triangles, $x''_p$ and $y''_p$ can be obtained as: 

\begin{equation}
x''_p=x''_t\cdot \frac{n}{|z''_t|}
\label{xproj}
\end{equation}

\begin{equation}
y''_p=y''_t\cdot \frac{n}{|z''_t|}
\label{yproj}
\end{equation}

In the image plane rectangle $E''F''G''H''$, the location of a projected point can be determined by its local 2D coordinates. First, the height $|E''H''|$ and width $|H''G''|$ are obtained using:

\begin{equation}
|E''H''|=2n \cdot tan{\left(\frac{\alpha}{2}\right)}
\end{equation}

\begin{equation}
|H''G''|=r|E''H''|
\end{equation}

Considering the bottom-left vertex $H''$ as the origin and utilizing the pixel as the unit length, we obtain the coordinates $(u, v)$ of projected point $P$ as:

\begin{equation}
u=w{\cdot}{\left(\frac{x''_p}{|H''G''|}{+}0.5\right)}=w{\cdot}{\left(\frac{x''_t}{2r|z''_t|tan{\left(\frac{\alpha}{2}\right)}}{+}0.5\right)}
\label{uproj}
\end{equation}

\begin{equation}
v=h{\cdot}{\left(\frac{y''_p}{|E''H''|}{+}0.5\right)}=h{\cdot}{\left(\frac{y''_t}{2|z''_t|tan{\left(\frac{\alpha}{2}\right)}}{+}0.5\right)}
\label{vproj}
\end{equation}

\noindent where $w$ and $h$ are the width and height of the camera image in pixels.

\subsubsection{Getting images from the virtual PTZ camera}
The virtual PTZ camera is designed to simulate as well as possible an off-the-shelf PTZ camera. First, it returns an image based on the current FOV as defined by $\theta_d$, $\phi_d$ and $\alpha$ (see Fig. \ref{fig:geometry1}b). Points in world coordinates are projected using (\ref{xproj}) and (\ref{yproj}) and expressed in image coordinates using (\ref{uproj}) and (\ref{vproj}). In our implementation, the virtual camera can provide images only when it is still. We preferred this option to artificially generating motion blur. Although debatable, previous works seem to agree on the fact that images with strong motion blur are not really usable \cite{Darvish2011}.

\subsubsection{Camera control}
\label{ipipeline}
The virtual PTZ camera also includes commands to change its orientation, either by using specific pan and tilt angles, or by recentering on a pixel position expressed in image coordinates. The zoom can also be simulated by changing the vertical FOV angle of the camera ($\alpha$). To reproduce the behavior of an actual PTZ camera, we consider that orientation changes are not instantaneous. Instead, the simulated camera pans and tilts based on the maximal angular speeds of a commercial PTZ camera (Sony SNC-RZ50N). This means that the image acquisition delay after a reorientation depends on the amplitude of camera's motion. To simulate this first type of delay (which we note $\tau_m$), frames are simply skipped in the video. 

A PTZ camera can pan and tilt to face a certain direction that the user wants. If the tracker finds an object of interest in the current image, a good way to follow it is to orient the camera on its current position (or its next predicted position), so that it stays near the center of the FOV. Inversing the ``forward'' pipeline of section \ref{pipeline}, we can compute the required camera direction vector by using the current 2D coordinates $(u,v)$ of the point to follow and known camera parameters (initial direction, FOV and output image size).

The geometric relations used to rotate the camera in order to center it on a point in the image plane is shown in Fig \ref{fig:geometry2}c, which is in world space. Originally, target $T$ is projected to point $P$ in the image plane $EFGH$, but $P$ is not at the center (D) of this rectangle. Then, the virtual camera is rotated both vertically and horizontally to the new direction $\vec{OD'}$, where the center $D'$ of image plane $E'F'G'H'$ is co-located with $\vec{OT}$. In this way, the target $T$ will be projected to $D'$ and placed at the center of the camera view. To obtain the required direction vector of the camera ($\vec{OD'}$), we can compute the direction of $\vec{OP'}$. In fact, both these vectors have the same direction as $\vec{OT}$. However, only $\vec{OT}$ is independent of the camera. For convenience, we describe the direction with the tilt angle $\theta_d$ and pan angle $\phi_d$ of $\vec{OT}$ shown in Fig. \ref{fig:geometry1}a. Hence, our task can be described as obtaining $\theta_d$ and $\phi_d$.

With the 2D coordinates $(u,v)$ of the projected point $P''$ in image space, the direction of the original point is obtained by reversing the projection process of section \ref{pipeline}. There are two steps to follow: 1) transform the image coordinates to camera coordinates using:
\begin{equation}
x''_p={\left({\frac{u}{w}}{-}0.5\right)}{\cdot}|H''G''|=2nr{\left({\frac{u}{w}}{-}0.5\right)}tan{\left(\frac{\alpha}{2}\right)}
\end{equation}
\begin{equation}
y''_p={\left({\frac{v}{h}}{-}0.5\right)}{\cdot}|E''H''|=2n{\left({\frac{v}{h}}{-}0.5\right)}tan{\left(\frac{\alpha}{2}\right)}
\end{equation}
\begin{equation}
z''_p=-n
\end{equation}
\noindent where $w$ and $h$ are the width and height of the image in pixels; and 2) transform the camera coordinates to world coordinates using:
\begin{equation}
\begin{bmatrix}
x_p\\
y_p\\
z_p
\end{bmatrix}=M^{-1}_{view}\cdot \begin{bmatrix}
x''_p\\
y''_p\\
z''_p
\end{bmatrix}
\end{equation}
\noindent Then, we can obtain $\theta_t$ and $\phi_t$ using:
\begin{equation}
\theta_t=arccos{\left(\frac{z_p}{\sqrt{x^2_p+y^2_p+z^2_p}}\right)}
\end{equation}
\begin{equation}
\phi_t=arctan{\left(\frac{y_p}{x_p}\right)}
\end{equation}

\subsubsection{Delay simulation}
A second type of delay, noted $\tau_p$, corresponds to the time required for a tracker to process a frame. We also consider a third type of delay, $\tau_c$, which is the communication delay over a network in the case of an IP PTZ camera. The user fixes this last delay. Note that all delays are simulated by skipping frames in the pre-recorded videos to mimic dropped frames.

Following a camera motion, the tracker will observe the scene again after a $\tau=\tau_m+\tau_p+\tau_c$ delay. Therefore, ideally, a tracker should try to minimize $\tau_p$ as much as possible and also try to predict the position of the target after the $\tau$ delay to make sure it stays in the FOV.

\subsection{Performance Evaluation}
Apart from the basic operations described in the last section, our PTZ camera framework also calculates four performance metrics to evaluate an actual tracker. Let $c_{GT}^t$ and $c_{PT}^t$ be the center locations of the ground-truth target and the predicted target (by the tracker) at time $t$, respectively, and $A_{GT}^t$ and $A_{PT}^t$ be the bounding boxes of the ground-truth target and the predicted target at time $t$, respectively, and $c_{FOV}^t$ be the location of the center of the image FOV at time $t$. These values are all expressed in image coordinates.
\begin{itemize}
\item $CLE$ (Center Location Error) at time $t$ is defined as
\begin{equation}
CLE^t=| c_{GT}^t- c_{PT}^t |
\end{equation}
It is invalid and assigned -1 if the target is out of the FOV. Its corresponding overall metric $CLE$ is the average of all valid $CLE^t$. This metric evaluates the quality of target localization.
\item $OR$ (Overlap Ratio) at time $t$ is defined as
\begin{equation}
OR^t= \frac{A_{GT}^t \cap A_{PT}^t} {A_{GT}^t \cup A_{PT}^t}
\end{equation}
Its corresponding overall metric $OR$ is the average of all $OR^t$. This metric also evaluates the quality of target localization.
\item $TCE$ (Target to Center Error) at time $t$ is defined as
\begin{equation}
TCE^t=| c_{FOV}^t- c_{GT}^t |
\end{equation}
It is invalid and assigned -1 if the target is out of the FOV. Its corresponding overall metric $TCE$ is the average of all valid $TCE^t$. This metric evaluates the quality of the camera control.
\item TF (Track Fragmentation) at time $t$ is defined as
\begin{equation}
TF^t= \left\{ \begin{array}{ll}
1 & \mbox{if $CLE^t$ is invalid}\\
0 & \mbox{otherwise}
\end{array}
\right. 
\end{equation}
This metric indicates if the target is inside or outside the FOV. Its corresponding overall metric $TF$ is the sum of all $TF^t$ divided by the number of processed frames. This metric also evaluates the quality of the camera control.
\end{itemize} 

\subsection{Spherical Panoramic Scenarios}
\label{scenario_definition}
Three spherical video sequences were captured in two indoor environments with 4 or 5 randomly moving persons. The videos contain a total number of 3,179 panoramic frames recorded at a frame rate of 16 fps (the maximum frame rate of the Ladybug 3). The first video was captured in a laboratory room cluttered with desks, chairs, posters and technical video equipment in the background. The Ladybug 3 camera was mounted on a tripod and placed in the center of the room. The two other spherical videos were recorded in the middle of a large atrium within a building with glass walls causing uneven illumination conditions. Our complete dataset includes 36 manually annotated tracking sequences extracted from the three initial spherical video sequences. The length of each tracking sequence varies from a few seconds to one or two minutes. For each of the 36 annotated sequences, the tracked target is one of the following: the full body of a moving person, a torso, a head, or an object carried by a person.  

\begin{figure}
    \centering
    \includegraphics [width=\columnwidth] {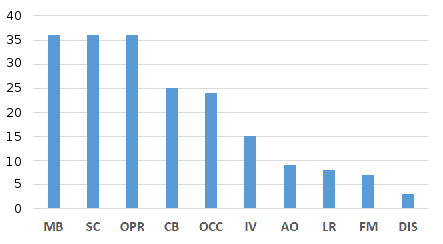}
    \caption{Difficulty distribution over the whole dataset.}
    \label{fig:dataset_distribution}
\end{figure}

\begin{figure}
    \centering
    \includegraphics [width=\columnwidth] {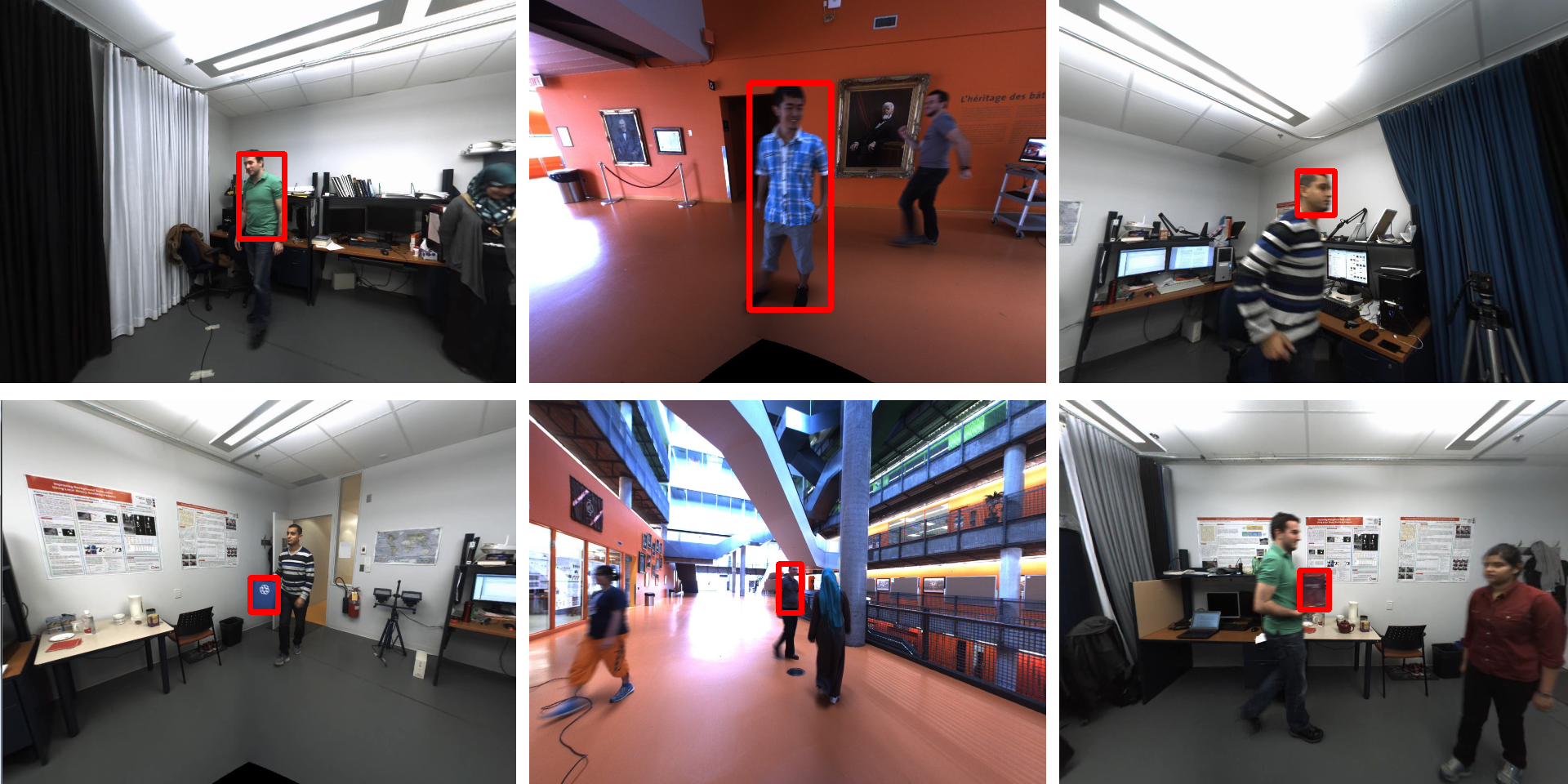}
    \caption{Examples of tracked objects in the proposed dataset.}
    \label{fig:dataset_examples}
\end{figure}

In real world scenarios, many perturbation factors can affect tracking performance. For our dataset, we used the difficulty categorization proposed in \cite{wu2013}. Three tracking difficulties are present in all our sequences: Motion Blur (MB), Scale Change (SC), and Out-of-Plane Rotation (OPR). Moreover, we defined subsets of videos corresponding to other perturbation factors: Fast Motion (FM), Cluttered Background (CB), Illumination Variation (IV), Low Resolution (LR), Occlusion (OCC), presence of Distractors (DIS), and Articulated Objects (AO). The histogram of Fig. \ref{fig:dataset_distribution} illustrates the difficulty distribution in our dataset. Note that one tracking sequence may include multiple difficulties. Fig. \ref{fig:dataset_examples} shows examples of targets that are tracked in our sequences.

\subsection{Ground-Truth}
In order to evaluate actual tracking methods, getting accurate tracking ground-truth (GT) sequences for our scenarios is important. The GT annotations for a PTZ camera video are very different from those of a fixed camera video. A traditional camera with a fixed view frustum simply requires a sequence of bounding boxes for each tracked object, which can all be defined by width, height and 2D center position. A PTZ camera GT is much more complex: the camera may have different FOV angles and output image sizes to observe the same target. It is necessary to make the GT applicable to all possible observation configurations. 

\subsubsection{Basic ground-truth}
First, we define the ``basic'' GT as the GT that is manually annotated for a chosen camera position. The actual GT required for evaluation in any other situation can be obtained by directly transforming the basic GT.

\begin{figure}
    \centering
    \subfloat[\label{fig:GTa}]{\includegraphics[width=0.47\linewidth]{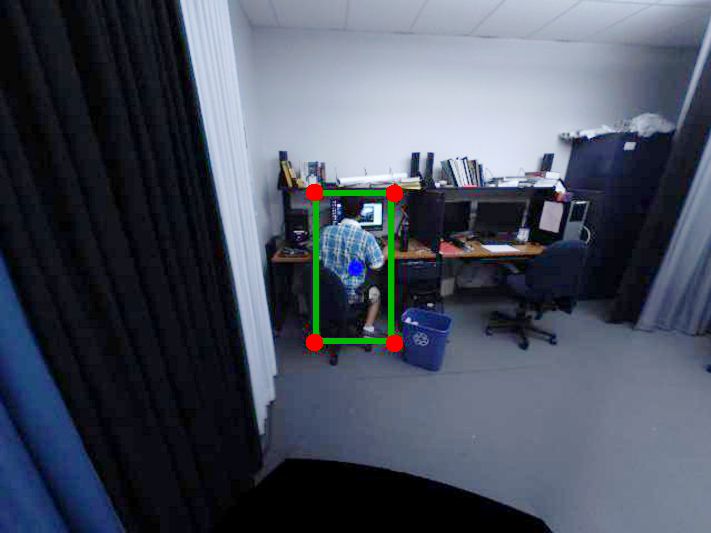}}
    \subfloat[\label{fig:GTb}]{\includegraphics[width=0.47\linewidth]{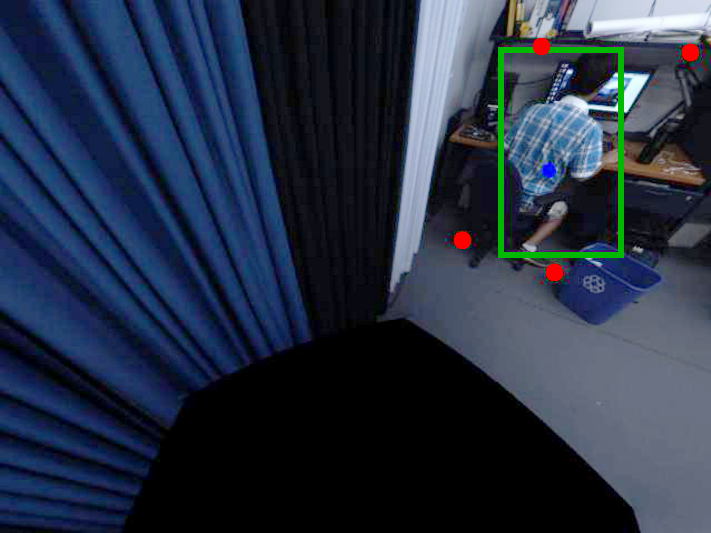}}
    \caption{Transformation of bounding boxes. (a) shows an example of basic GT (the target is centered in the image), and (b) shows how the four vertices of the original bounding box are mapped to four points in the new FOV before being rectified to make a rectangle.}
    \label{fig:GT} 
\end{figure}

For the basic GT, during the annotation phase, the target is always in the middle of the image, as shown as Fig. \ref{fig:GT}a. Four values are recorded for each frame of the video, which can be transformed to different target points and bounding boxes according to the virtual camera's viewing direction, FOV and output image size. Two of these four values are related to the current orientation of the camera: they are the pan and tilt angles. The other two values are simply the width and height of the target's bounding box. While collecting basic GT, the other parameters of the camera are fixed but they must also be recorded since they are used in the evaluation phase. These parameters are the camera's vertical FOV angle and its output image width and height.

\subsubsection{Ground-truth adjustment for current FOV}
In the basic GT, the target center point is always at the center of the camera's output image; as a result, the direction to the target ($\vec{OT}$) is the same as the camera direction ($\vec{OD}$) (a similar situation is demonstrated in Fig. \ref{fig:geometry2}c). 

In order to obtain the 2D coordinates of the target on the image plane of a camera with different parameters, we can find its 3D coordinates in the world space by using its direction. It is first transformed to camera coordinates, and then to image coordinates using the equations of section \ref{pipeline}.

The transformation of bounding boxes is more complicated. Not only can their position shift, but their size, width/height ratio, and shape can vary. We use the four vertices of the rectangle to represent the bounding box, and compute their 2D coordinates on the image plane. Using the inverted pipeline of section \ref{ipipeline}, their directions (consisting of horizontal and vertical angles) are computed. Then, using the configuration of the tracker's virtual camera, these four directions are projected back to the tracker's image plane with the forward pipeline (section \ref{pipeline}). Due to distortion (the bounding box is not viewed in the same plane as originally drawn), the four projected points are unlikely to make up a rectangle, as shown in Fig. \ref{fig:GT}b. They are thus rectified by projecting the pairwise means of their 2D coordinates on the horizontal and vertical axes of the image space. Then, they are connected on these same axes to form an axis-aligned bounding box. A rectified bounding box is also shown in Fig. \ref{fig:GT}b.

\section{Baseline Evaluation Results}

\begin{table}
\centering
\begin{tabular}{| c c c c c |}
\hline
         & $\tau_c=0 $ & $\tau_c=\sfrac{1}{8}$ & $\tau_c=\sfrac{1}{4}$ & $\tau_c=\sfrac{1}{2}$\\
\hline\hline
               \textit{CB}  & 99.1 &107.1 &113.9 &139.5 \\
\rowcolor{Gray}\textit{OCC} & 86.7 & 96.6 &100.5 &117.8 \\
               \textit{IV}  &130.5 &143.9 &150.7 &152.6 \\
\rowcolor{Gray}\textit{AO}  &150.9 &148.0 &153.8 &141.5 \\
               \textit{LR}  &129.6 &133.3 &175.6 &136.2 \\
\rowcolor{Gray}\textit{FM}  & 80.5 &113.6 &151.6 &134.3 \\
               \textit{DIS} & 30.9 & 39.1 & 42.6 &158.6 \\
\hline\hline
      \textbf{full dataset} & 83.2 & 89.9 & 92.1 &123.8 \\
\hline
\end{tabular}
\vspace{-9pt}
\caption { Center Location Error (CLE) in pixels for Camshift with four different communication delays in seconds.}
\vspace{-8pt}
\label{tab:cle}
\end{table}
\begin{table}
\centering
\begin{tabular}{| c c c c c |}
\hline
         & $\tau_c=0 $ & $\tau_c=\sfrac{1}{8}$ & $\tau_c=\sfrac{1}{4}$ & $\tau_c=\sfrac{1}{2}$\\
\hline\hline
               \textit{CB}  & 97.2 &104.1 &111.3 &134.6 \\
\rowcolor{Gray}\textit{OCC} & 85.3 & 93.1 &100.5 &116.0 \\
               \textit{IV}  &128.4 &138.6 &143.4 &148.1 \\
\rowcolor{Gray}\textit{AO}  &146.9 &140.0 &144.0 &145.0 \\
               \textit{LR}  &127.4 &125.9 &161.0 &131.9 \\
\rowcolor{Gray}\textit{FM}  & 81.1 &105.8 &146.1 &132.5 \\
               \textit{DIS} & 33.0 & 44.6 & 54.2 &161.6 \\
\hline\hline
      \textbf{full dataset} & 81.9 & 88.3 & 93.5 &123.5 \\
\hline
\end{tabular}
\vspace{-9pt}
\caption { Target to Center Error (TCE) in pixels for Camshift with four different communication delays in seconds.}
\vspace{-8pt}
\label{tab:tce}
\end{table}
\begin{table}
\centering
\begin{tabular}{| c c c c c |}
\hline
         & $\tau_c=0 $ & $\tau_c=\sfrac{1}{8}$ & $\tau_c=\sfrac{1}{4}$ & $\tau_c=\sfrac{1}{2}$\\
\hline\hline
               \textit{CB}  & 0.260 & 0.234 & 0.197 & 0.121 \\
\rowcolor{Gray}\textit{OCC} & 0.323 & 0.303 & 0.259 & 0.228 \\
               \textit{IV}  & 0.207 & 0.169 & 0.122 & 0.030 \\
\rowcolor{Gray}\textit{AO}  & 0.239 & 0.183 & 0.147 & 0.118 \\
               \textit{LR}  & 0.274 & 0.203 & 0.110 & 0.064 \\
\rowcolor{Gray}\textit{FM}  & 0.327 & 0.263 & 0.257 & 0.156 \\
               \textit{DIS} & 0.405 & 0.401 & 0.415 & 0.110 \\
\hline\hline
      \textbf{full dataset} & 0.317 & 0.298 & 0.273 & 0.195 \\
\hline
\end{tabular}
\vspace{-9pt}
\caption { Overlap Ratio (OR) for Camshift with four different communication delays in seconds.}
\vspace{-8pt}
\label{tab:or}
\end{table}
\begin{table}
\centering
\begin{tabular}{| c c c c c |}
\hline
         & $\tau_c=0 $ & $\tau_c=\sfrac{1}{8}$ & $\tau_c=\sfrac{1}{4}$ & $\tau_c=\sfrac{1}{2}$\\
\hline\hline
               \textit{CB}  & 0.470 & 0.467 & 0.498 & 0.581 \\
\rowcolor{Gray}\textit{OCC} & 0.482 & 0.481 & 0.446 & 0.491 \\
               \textit{IV}  & 0.562 & 0.592 & 0.613 & 0.687 \\
\rowcolor{Gray}\textit{AO}  & 0.522 & 0.515 & 0.618 & 0.541 \\
               \textit{LR}  & 0.466 & 0.543 & 0.592 & 0.668 \\
\rowcolor{Gray}\textit{FM}  & 0.543 & 0.602 & 0.467 & 0.595 \\
               \textit{DIS} & 0.188 & 0.282 & 0.307 & 0.735 \\
\hline\hline
      \textbf{full dataset} & 0.440 & 0.442 & 0.405 & 0.520 \\
\hline
\end{tabular}
\vspace{-9pt}
\caption { Track Fragmentation (TF) for Camshift with four different communication delays in seconds.}
\vspace{-8pt}
\label{tab:tf}
\end{table}

In order to provide baseline tracking results, we used our virtual camera framework to evaluate a simple PTZ tracker based on the well-known Camshift algorithm \cite{bradski1998real}. Looking back at the two general PTZ tracking design families described in section \ref{intro}, we can classify this tracker as part of the first family (i.e. fast, but not very robust). As such, we used the typical camera control strategy of this design family, meaning that the camera FOV is continuously recentered at the target's previous location.

We tested the Camshift tracker on the proposed 36 sequences at a 640x480 resolution with a 90$^{\circ}$ vertical FOV angle, using the categorized difficulties defined in section \ref{scenario_definition}. Tables \ref{tab:cle}, \ref{tab:tce}, \ref{tab:or}, and \ref{tab:tf} present the results of Center Location Error (CLE), Target to Center Error (TCE), Overlap Ratio (OR), and Track Fragmentation (TF), respectively. In all these experiments, we simulated the camera motion delay $\tau_m$ of the commercial PTZ camera Sony SNC-RZ50N based on its maximal angular speed of 300$^{\circ}$/s. We considered the processing delay of the tracker ($\tau_p$) as the actual execution time of Camshift for each frame. However, since the Camshift tracker can typically process more than 16 fps (we ran it on an Intel i5 3570 CPU at 3.4 GHz), its processing delay can be considered null ($\tau_p=0$), as it will not cause a significant number of frames to be skipped. On the other hand, we evaluated this tracker using four different communication delays ($\tau_c$) as shown in the tables. Note that in these tables, the Motion Blur (MB), Scale Change (SC), and Out-of-Plane Rotation (OPR) difficulties are not included because all the sequences of our dataset contain them. As a result, studying these difficulties is equivalent to studying the full dataset. Also, recall that sequences are not exclusive to any difficulty category. For instance, sequences in the Occlusion (OCC) category present some form of occlusion but may also present illumination variations and thus be part of the Illumination Variations (IV) category.

From our results, we can see that Camshift does not offer very good performance for the challenges present in typical PTZ tracking problems. For example, localization errors reported by the CLE and TCE metrics exceed 80 pixels for all but one difficulty, and Track Fragmentation (TF) is almost always above 0.450. While Camshift's histogram-based approach is effective for short sequences with no occlusions, it was unable to track targets with no vivid colors or with an appearance that was similar to the background, which make up a good proportion of our test sequences. Furthermore, in all sequences that provide an initialization bounding box with visible background, Camshift rapidly dropped its target and started drifting through the entire scene randomly. However, in sequences where the target is brightly colored, only suffers from partial occlusions, and does not resemble the background, Camshift took advantage of its high processing speed to keep track of the target. Overall, these baseline results show the usability of our framework and demonstrate that tracking and controlling the virtual PTZ successfully on our dataset is not trivial. More sophisticated tracking algorithms are required to solve its challenges.

While it is hard to directly compare the proposed test subsets due to their varying sizes and overlaps and the nature of their targets, we note that the scores obtained for all four metrics in the Distractors (DIS) category are generally better than those of any other category. In DIS, all targets are human heads, which are rather easy to track with a histogram-based method, as long as the background does not match skin color. The Illumination Variation (IV) category seemed to be the hardest to handle for Camshift; extreme contrast and camouflage problems due to intense light sources sometimes made tracking nearly impossible. The Low Resolution (LR) and Articulated Objects (AO) categories share multiple sequences with IV, which might explain their similar scores.

More generally, we can observe that increasing the communication delay ($\tau_c$) has a deep impact on the effectiveness of the tracking algorithm. While it could be expected that increasing this parameter's value would directly worsen tracking performances, interestingly, this is not always the case. In fact, for a handful of sequences presenting full occlusions, adding a communication delay sometimes helps the tracker by either completely eliminating these occlusions or by replacing them with partial occlusions. This is however uncommon. More typically, all metrics except Track Fragmentation (TF) show decreases in performance for each increment of $\tau_c$.

\section{Conclusion}

In this paper, we have proposed a new publicly available framework for the evaluation of PTZ tracking algorithms. It allows realistic experiments to be repeated in identical conditions. This framework simulates a PTZ camera that can pan, tilt, and zoom to observe different parts of a scene constructed using pre-recorded spherical panoramic videos. It also considers various types of delays and limitations of commercial PTZ cameras to provide a faithful reproduction of real tracking experiments.

We provide a total of 36 annotated tracking sequences along with our PTZ framework, which sum up to over 16,000 bounding boxes. Unlike the ground-truth used for fixed-camera tracking, these bounding boxes were defined using a spherical coordinate system and can be used to evaluate tracking performance under different camera configurations. To provide baseline results for our framework, we tested the Camshift algorithm on these 36 sequences. The four metrics we use to evaluate PTZ tracking performance indicate that Camshift is generally unable to handle the challenges present in typical PTZ tracking scenarios. We are confident that tracking methods specifically designed for PTZ scenarios can overcome the realistic difficulties present in our dataset.

{\small
    \bibliographystyle{ieee}
    \bibliography{references}
}

\end{document}